\pdfoutput=1
\documentclass[11pt,a4paper]{article}
\usepackage[final]{acl}
\usepackage{times}
\usepackage{latexsym}

\usepackage{amsmath}
\usepackage{amssymb}
\usepackage[normalem]{ulem}
\usepackage{graphicx}
\usepackage{caption}
\usepackage{booktabs}
\usepackage{multirow}
\usepackage{xcolor}
\usepackage{lipsum}
\usepackage{placeins}
\usepackage{bbm}
\usepackage{makecell}
\usepackage{ucs}
\usepackage[utf8]{inputenc}
\usepackage{tikz}
\usepackage{utfsym}
\usepackage{wrapfig,lipsum,booktabs}
\usepackage{microtype}
\usepackage{pifont}
\usepackage{enumitem}
\usepackage[colorinlistoftodos,prependcaption,textsize=tiny]{todonotes}
\usepackage{xspace}
\definecolor{orangycolor}{HTML}{CC4125}
\definecolor{greensea}{HTML}{16A085}
\definecolor{lb}{rgb}{.90,.95,1}
\definecolor{ly}{rgb}{0.98, 0.91, 0.71}
\definecolor{lo}{rgb}{1.0, 0.6, 0.4}
\definecolor{lg}{rgb}{0.67, 0.88, 0.69}
\usepackage{xargs} % commandx
\newcommandx{\siva}[2][1=]{\todo[linecolor=red,backgroundcolor=red!10,bordercolor=red,#1]{SR: #2}\xspace}
\newcommandx{\zichao}[2][1=]{\todo[linecolor=green,backgroundcolor=green!10,bordercolor=green,#1]{ZL: #2}\xspace}

\usepackage[export]{adjustbox}

\newcommand{\feed}{\textsc{FeedbackQA}\xspace}
\newcommand{\exprate}{\textsc{ExplainRate}\xspace}

\newcommand{\feedranker}{\textsc{FeedbackReranker}\xspace}
\newcommand{\vanillaranker}{\textsc{VanillaReranker}\xspace}
\newcommand{\combinedranker}{\textsc{CombinedReranker}\xspace}

\newcommand{\rate}{\textsc{RateOnly}\xspace}

\usepackage{cleveref}
\Crefformat{equation}{(#2#1#3)}

\newcommand \ignore[1]{}

\usepackage{color,soul}
\newcommand{\hlc}[2][yellow]{ {\sethlcolor{#1} \hl{#2}} }
\title{Using Interactive Feedback to Improve the  Accuracy and Explainability of Question Answering Systems Post-Deployment}
\author{
    Zichao Li\textsuperscript{\rm 1},
    ~~Prakhar Sharma\textsuperscript{\rm 2},
    ~~Xing Han Lu\textsuperscript{\rm 1},
    ~~Jackie C.K. Cheung\textsuperscript{\rm 1},
    ~~Siva Reddy\textsuperscript{\rm 1}\\

    \textsuperscript{\rm 1}Mila, McGill University~ \\
    \textsuperscript{\rm 2}University of California, Los Angeles \\
    
    \texttt{zichao.li@mila.quebec}\\
}

\date{}

\begin{document}
\maketitle
\begin{abstract}
Most research on question answering focuses on the pre-deployment stage; i.e., building an accurate model for deployment.
In this paper, we ask the question: Can we improve QA systems further \emph{post-}deployment based on user interactions? 
We focus on two kinds of improvements: 1) improving the QA system's performance itself, and 2) providing the model with the ability to explain the correctness or incorrectness of an answer.
We collect a retrieval-based QA dataset, \feed{}, which contains interactive feedback from users. We collect this dataset by deploying a base QA system to crowdworkers who then engage with the system and provide feedback on the quality of its answers.
The feedback contains both structured ratings and unstructured natural language explanations.
We train a neural model with this feedback data that can generate explanations and re-score answer candidates. We show that feedback data not only improves the accuracy of the deployed QA system but also other stronger non-deployed systems. 
The generated explanations also help users make informed decisions about the correctness of answers.\footnote{Project page: \href{https://mcgill-nlp.github.io/feedbackqa/}{https://mcgill-nlp.github.io/feedbackqa/}}
%At present, they are uploaded to google drive via an anonymized account: \url{https://drive.google.com/drive/folders/1G3lym-pbOzwrMuSeV_Shx3PFX25q_QGl?usp=sharing}}
% We release the dataset to the community to foster research in improving QA systems from user interactions.

\end{abstract}
\section{Introduction}\label{sec:intro}

Much of the recent excitement in question answering (QA) is in building high-performing models with carefully curated training datasets.
Datasets like SQuAD \citep{rajpurkar2016squad}, NaturalQuestions \citep{kwiatkowski2019natural} and CoQA \citep{reddy2019coqa} have enabled rapid progress in this area.
%This progress has been further enabled by the advent of large pre-trained attention-based models \citep{devlin_bert_2019} which make use of large amounts of unsupervised data.
% , and more recently retrieval-augmented language models \citep{guu2019realm}.
%These QA models are also becoming the backbone for major search engines like Google, serving billions of users.\footnote{\href {https://blog.google/products/search/search-language-understanding-bert/}{Google blog: Understanding searches better than ever before}}
Most existing work focuses on the pre-deployment stage; i.e., training the best QA model before it is released to users. However, this stage is only one stage in the potential lifecycle of a QA system. 

In particular, an untapped resource is the large amounts of user interaction data produced after the initial deployment of the system. Gathering this data should in practice be relatively cheap, since users genuinely engage with QA systems (such as Google) for information needs and may provide feedback to improve their results.\footnote{Google and Bing collect such data through "Feedback" button located at the bottom of search results.}

Exploiting this kind of user interaction data presents new research challenges, since they typically consist of a variety of weak signals. For example, user clicks could indicate answer usefulness~\cite{Joachims2002opt}, users could give structured feedback in the form of ratings to indicate the usefulness~\cite{stiennon2020feedback4sum}, or they could give unstructured feedback in natural language explanations on why an answer is correct or incorrect. %With the rise of AI assistants, the natural language feedback could soon become more prevalent. 
User clicks have been widely studied in the field of information retrieval \cite{Joachims2002opt}. 
Here we study the usefulness of \textit{interactive feedback} in the form of ratings and natural language explanations.

\begin{figure*}
%\hspace{-3em}
%\includegraphics[trim=0em 5em 0em 9em,clip=true,width=1.3\columnwidth]{figures/FeedbackQA}
\centering
\includegraphics[width=\textwidth]{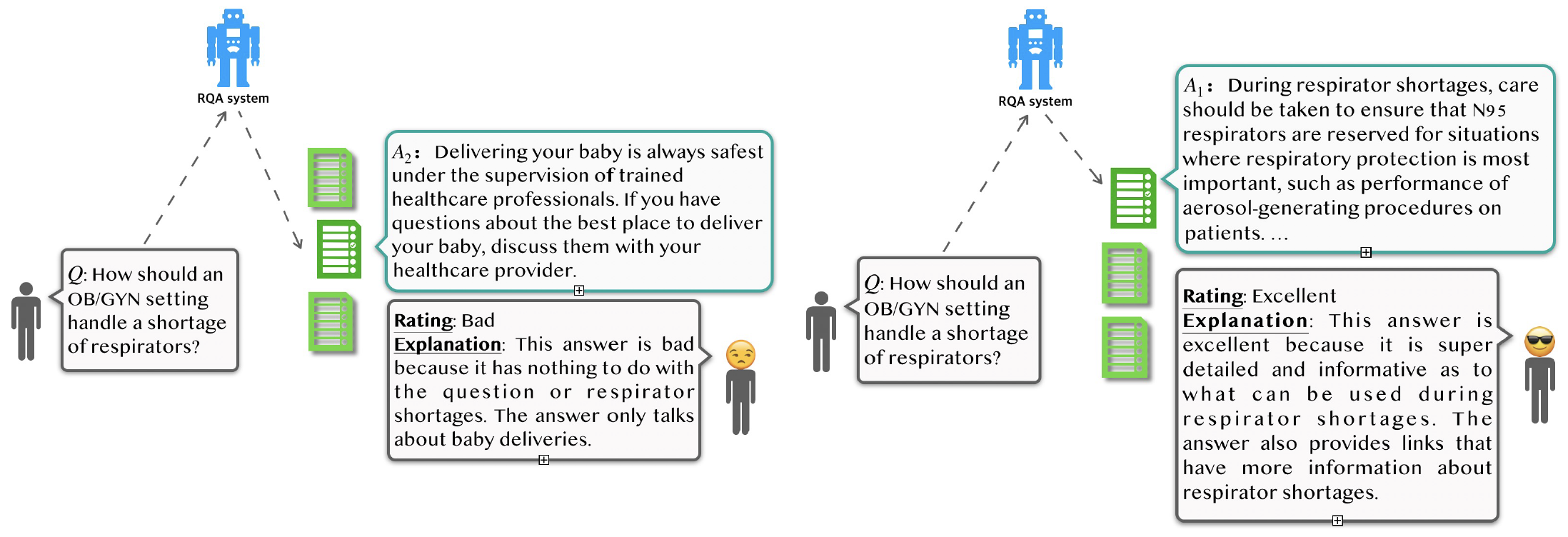}
\caption{Users interact with the deployed QA model and give feedback. Feedback contains a rating (\textit{bad, good, could be improved, excellent}) and a natural language explanation.}
\label{fig:feedbackqa}
\end{figure*}

Whilst there are different variants of QA tasks, this paper focuses primarily on retrieval-based QA (RQA; \citealt{chen2017drqa,lee2019latent}).
Given a question and a set of candidate answer passages, a model is trained to rank the correct answer passage the highest.
In practice, when such a system is deployed, an user may engage with the system and provide feedback about the quality of the answers.
Such feedback is called interactive feedback.
% After the user reads the returned answer, he or she may find the passage useful or useless. If possible, the user could leave feedback regarding the usefulness and quality of the answer. The feedback could be a rating label and/or a natural language explanation. Unfortunately, there is a lack of an open RQA dataset containing such interactive feedback. 
Due to the lack of a dataset containing interactive feedback for RQA, we create \feed{}.

%a large-scale English QA dataset containing interactive feedback of the form user ratings (structured) and natural language explanations (unstructured) about the correctness of an answer.
% To make our dataset practically useful, we work with data from public health agencies for Covid-19 pandemic.
% The base model for \feed is built on $27{,}722$ questions and $2{,}951$ passages from various agencies.
% We collected $9{,}434$ interactive feedback data samples using the base model.

\feed is a large-scale English QA dataset containing interactive feedback in two forms: user ratings (structured) and natural language explanations (unstructured) about the correctness of an answer.
Figure~\ref{fig:feedbackqa} shows an example from \feed{}.
The dataset construction has two stages: 
We first train a RQA model on the questions and passages, then deploy it on a crowdsourcing platform.
Next, crowdworkers engage with this system and provide interactive feedback. 
To make our dataset practically useful, we focus on question answering on public health agencies for the Covid-19 pandemic.
The base model for \feed is built on $28$k questions and $3$k passages from various agencies.
We collect $9$k interactive feedback data samples for the base model.

% The dataset construction has two stages: We first train a strong RQA model on the questions and passages data, then deploy it on a crowdsourcing platform, where we collect interactive feedback on the RQA model. 
% Whilst the explanation data in existing QA datasets is either in a structured form or in natural language, we ask the crowdsourcing workers to give feedback of two types: a rating label and a natural language explanation regarding the answer quality. 

We investigate the usefulness of the feedback for improving the RQA system in terms of two aspects: answer accuracy and explainability. 
Specifically, we are motivated by two questions: 
1) Can we improve the answer accuracy of RQA models by learning from the interactive feedback? 
and 2) Can we learn to generate explanations that help humans to discern correct and incorrect answers?  
% explainability, by giving it the ability to explain both the strengths and/or weaknesses of its predictions, to inform the user while they read the passage?
% \end{enumerate}
% We argue that the second aspect is particularly significant in domains such as health and education, where users have higher expectations about the trustworthiness of the system they interact with.
%\siva{Highlight the importance of explainability in domains like health -- increases trust}
%\siva{Talk about how this dataset differs from existing datasets. Refer to \Cref{tab:data_char}}

% We investigate methods that can make use of interactive feedback to improve the base QA model's performance.

To address these questions, we use feedback data to train models that rerank the original answers as well as provide an explanation for the answers.
% improve the  improa model called \exprate, which generates an explanation for a question-answer pair, which is used as augmented context for the final selection of answer candidates.
Our experiments show that this approach not only improves the accuracy of the base QA model for which feedback is collected but also other strong models for which feedback data is not collected. 
Moreover, we conduct human evaluations to verify the usefulness of explanations and find that the generated natural language explanations help users make informed and accurate decisions on accepting or rejecting answer candidates.

% Our experimental results show that 

% Our main contributions are : 1) a large-scale dataset containing interactive feedback for QA; 2) methods that leverage interactive feedback to improve QA performance; and 3) methods to improve the explainability of QA models. 

\ignore{
\begin{figure*}
    \centering
    \includegraphics[width=\linewidth]{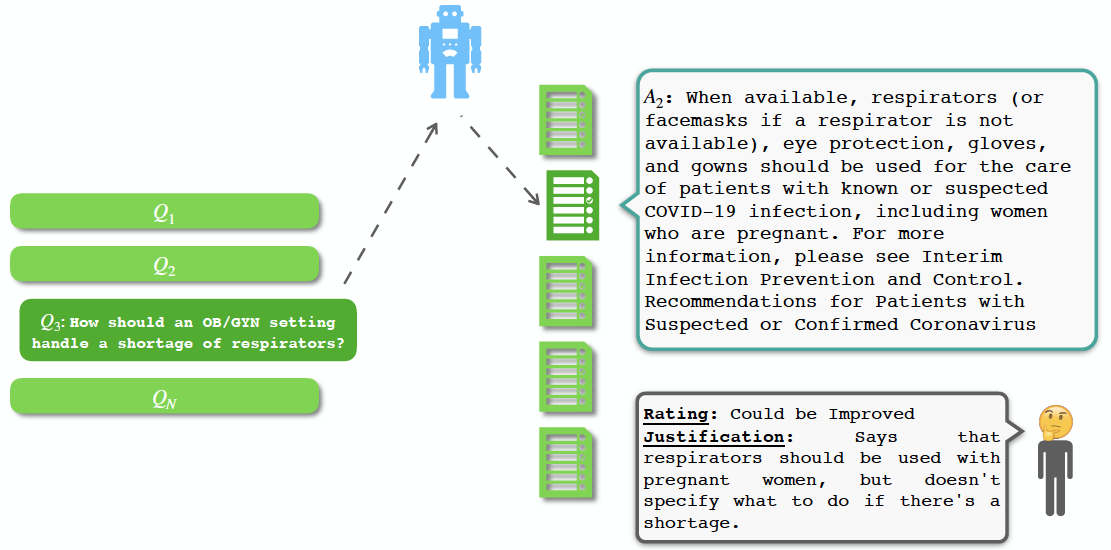}
    \caption{Captions}
    \label{fig:my_label}
\end{figure*}
}

% With the newly crawled data and interactive feedback, we pose two challenges to the communities of question answering and information retrieval: (1) how to endow the QA system to generate fluent and faithful explanation to rationalize its prediction; (2) how can the QA system improve by learning from the human explanation. 

% To investigate these questions preliminarily, we benchmark three base tasks, including retrieval-based question answering, rating prediction and explanation generation. Furthermore we incorporate them towards explainable retrieval-based question answering. We investigate the current state-of-the-art approach to explainable natural language understanding, generating explanation as augmented context for final prediction. We found that this approach failed in our target task. Thereby, we design a new approach based on multi-task learning, and the empirical experiment results shows that it is promising. It can be adopted as a strong benchmark for future studies.

Our contributions are as follows:
\begin{enumerate}[nolistsep]
\item We create the first retrieval-based QA dataset containing interactive feedback.
\item We demonstrate a simple method of using the feedback data to increase the accuracy and explainability of RQA systems. 
\item We show that the feedback data not only improve the deployed model but also a stronger non-deployed model.
\end{enumerate}

\section{\feed Dataset}
\label{sec:dataset}
Recently, there have been efforts to collect feedback data in the form of explanations for natural language understanding tasks~(\citealt{camburu_e-snli_2018,rajani2019explain}, \textit{inter alia}). 
These contain explanations only for ground-truth predictions for a given input sampled from the training data without any user-system interaction.
Instead, we collect user feedback after deploying a RQA system thereby collecting feedback for both correct and incorrect predictions.
% For retrieval-based QA (RQA), we argue that it is especially important to collect explanations for both positive and negative answer passages, as even negative passages may contain partially correct or relevant information.  
% The feedback for negative examples generated by the base model is also useful for model training and development. 
% The feedback is interactive in the sense that the negative examples come from a deployed system.
\Cref{tab:data_char} presents a comprehensive comparison of \feed{} and existing natural language understanding (NLU) datasets with explanation data.
%The existing NLU datasets with explanation only contain explanation for ground-truth. However, the RQA involves selection from a set of candidates. And thus the model also needs to know why the negative candidate is not the valid answer. However, it is too expensive to ask the human workers to explain every single pairs of question and potential candidates. Instead, we select a small set of negative examples to label with a well-trained QA model.
\begin{table*}[htp]
    \centering
    \resizebox{0.86\textwidth}{!}{
    \begin{tabular}{lrrrrrr}
        \toprule
         Datasets & Task &  Feedback & Interactive & %\makecell{Feedback for \\positive examples} & 
         Feedback for\\ 
         & & Type & Feedback & incorrect predictions \\
         \midrule
         e-SNLI~\cite{camburu_e-snli_2018} & NLI & Free-form  & \ding{55} &  \ding{55}\\
         % NILE~\cite{kumar_nile_2020} & NLI & Free-form  & \ding{55} & \ding{55} \\
         CoS-E~\cite{rajani2019explain} & Commonsense QA & Free-form  & \ding{55} &   \ding{55} \\
         LIAR-PLUS~\cite{alhindi2018liarplus} & Fact checking & Free-form & \ding{55} & \ding{55}\\
         QED~\cite{lamm2020qed} & Reading comprehension & Structured  & \ding{55} &  \ding{55} \\
         % \textsc{Break}~\cite{wolfson2020break} & Multi-hop QA  & Structured  & \ding{55} &  \ding{55} \\
         NExT~\cite{wang2019next} & Text classification & Structured  & \ding{55}  &\ding{55} \\
         \midrule
         \feed{} & Retrieval-based QA & Structured  & \ding{51} & \ding{51} \\
         & & \& Free-form \\
         \bottomrule
    \end{tabular}
    }
    \caption{Comparison of \feed{} with existing NLU datasets containing feedback in the form of structured representations (according to a schema) or  natural language explanations (free-form).}
    \label{tab:data_char}
\end{table*}
\vspace{-6pt}

\subsection{Dataset collection}
In order to collect post-deployment feedback as in a real-world setting, we divide the data collection into two stages: pre-deployment (of a RQA model) and post-deployment.
\vspace{-5pt}
\paragraph{Stage 1: Pre-deployment of a QA system}

We scrape Covid-19-related content from the official websites of  \href{https://www.who.int/emergencies/diseases/novel-coronavirus-2019/question-and-answers-hub}{WHO}, \href{https://www.cdc.gov/coronavirus/2019-nCoV/index.html}{US Government}, \href{https://www.gov.uk/coronavirus}{UK Government}, \href{https://www.quebec.ca/en/health/health-issues/a-z/2019-coronavirus}{Canadian government},\footnote{We focus on the Province of Quebec} and \href{https://www.health.gov.au/}{Australian government}.
We extract the questions and answer passages in the FAQ section. 
To scale up the dataset, we additionally clean the scraped pages and extract additional passages for which we curate corresponding questions using crowdsourcing as if users were asking questions.
% Then, we recruit crowd-source workers at the Amazon Mechanical Turk (AMT) platform to write questions that could have the given passage as their answers.
% whose answer is almost identical to the given passage. 
We present details on this annotation process in \Cref{sec:appendix-data-collect}.
We use this dataset to train a base RQA model for each source separately and deploy them.
For the base model, we use a BERT-based dense retriever \cite{karpukhin2020dense} combined with Poly-encoder \cite{miller2017parlai} (more details are in \Cref{sec:baseline-models}). 
% The details of Poly-encoder is introduced in . We fine-tune their pre-trained model on our dataset. 
\vspace{-5pt}
\paragraph{Stage 2: Post-deployment of a QA system}
Since each domain has several hundred passages (\Cref{tab:data_stat}), it is hard for a crowdworker to ask questions that cover a range of topics in each source. We thus collect questions for individual passages beforehand similar to Stage 1 and use these as interactive questions.
% We train a state-of-the-art RQA model proposed by ~\citet{humeau_poly-encoders_2020} on the new dataset, for use in the process of feedback collection.
% Since the set of answer candidates is large, it is hard for a crowdsourcing worker to produce questions that covered the range of topics.
% To circumvent this, given a passage, we collect questions beforehand and use these as interactive questions.
The question and top-2 predictions of the model are shown to the user and they give feedback for each question-answer pair. 
The collected feedback consists of a rating, selected from \textit{excellent, good, could be improved, bad}, and a natural language explanation elaborating on the strengths and/or weaknesses of the answer. 
For each QA pair, we elicit feedback from three different workers. 
We adopted additional strategies to ensure the quality of the feedback data, the details of which are available in \Cref{sec:appendix-feedback}. 
The resulting dataset statistics are shown in \Cref{tab:data_stat}.
In order to test whether interactive feedback also helps in out-of-distribution settings, we did not collect feedback for one of the domains (Canada).
%\paragraph{Dataset splits} We split the collected data into five sub-domains, which are categorized by their respective data sources: \textit{US}, \textit{Canada}, \textit{WHO}, \textit{UK}, and \textit{Australia}. Within each sub-domain, there are two test sets, which are in-domain and out-of-domain respectively. Concretely, for each sub-domain, about 10\% of the passages and their associated questions are identified for the out-of-domain test set, while others are split between training, validation, and an in-domain test set.    %10\% of the passages were selected randomly for each dataset and all the associated questions to those passages were added to their respective test set.
%Apart from this, 10\% of the original total collected questions for each dataset were randomly taken out of the remaining set and were added to the test set.
%Now the remaining 80\%(approx.) is further randomly split into 70\% train + dev(80\% and 20\%) and 30\% for our feedback collection. 
%Table \ref{tab:data_stat} shows the statistics of \feed{}.

\begin{table}[]
    \centering
    \resizebox{\linewidth}{!}{
    \begin{tabular}{lrrrr}
        \toprule
         & \#Passages
         & \#Questions
          & \#Feedback 
          &\\ \midrule
         Australia & 584 & 1783 & 2264 \\
         Canada & 587 & 8844 & /  \\ 
         UK & 956 & 2874 & 3668 \\
         US & 598 & 13533 & 2628 \\ %\hline
         WHO & 226 & 688 & 874\\
         \cmidrule{1-5}
         Overall & 2951 & 27722 & 9434 \\
         \bottomrule
    \end{tabular}
    }
    \caption{Number of samples in different domains of \feed{}. We split the data into train/validation/test sets in the ratio of $0.7:0.1:0.2$.}
    \label{tab:data_stat}
    \vspace{-\baselineskip}
\end{table}
\vspace{-8pt}

\subsection{\feed analysis}\label{sec:data-analysis}
% What do users talk about in their explanation?
\Cref{tab:exp-examples} shows examples of the feedback data, including both ratings and explanations.
We find that explanations typically contain review-style text indicating the quality of the answer, or statements summarizing which parts are correct and why.
Therefore, we analyze a sample of explanations using the following schema:
% In order to gain more insight into this, we analyze explanations along the following two components:

\noindent \textbf{Review} Several explanations start with a generic review such as \textit{This directly answers the question} or \textit{It is irrelevant to the question.} %This can be regarded as a direct indication of a user's sentiment towards the answer candidates. 
Sometimes users also highlight aspects of the answer that are good or can be improved. 
For instance, \textit{... could improve grammatically ...} suggests that the answer could be improved in terms of writing.

%\vspace{-8pt}
\vspace{-2pt}
% The majority of content consists of a summary of information presented in the question and answer candidates. 
\noindent \textbf{Summary of useful content} refers to the part of answer that actually answers the question;\\ 
\noindent \textbf{Summary of irrelevant content} points to the information that is not useful for the answer, such as off-topic or addressing incorrect aspects; \\
\noindent \textbf{Summary of missing content} points the information the answer fails to cover.
\vspace{-2pt}

We randomly sample 100 explanations and annotate them. 
\Cref{fig:my_label} shows the distribution of the types present in explanations for each rating label.
All explanations usually contain some review type information.
% Review type content is evenly distributed across all rating labels.
Whereas explanations for answers labeled as excellent or acceptable predominantly indicate the parts of the answer that are useful.
The explanations for answers that can be improved indicate parts that are useful, wrong or missing.
Whereas bad answers often receive explanations that highlight parts that are incorrect or missing as expected.
% As for the summary of question/answer, the explanation for high rating tends to contain only summary of useful content, whilst the ones for lower rating more likely contains summary of irrelevant and missing content.
% Half of the natural language feedback talks about the content of question and answer. %This shows that the feedback data in \feed is of high quality. 
\vspace{-2pt}
\begin{figure}
    \centering
    \includegraphics[scale=0.43]{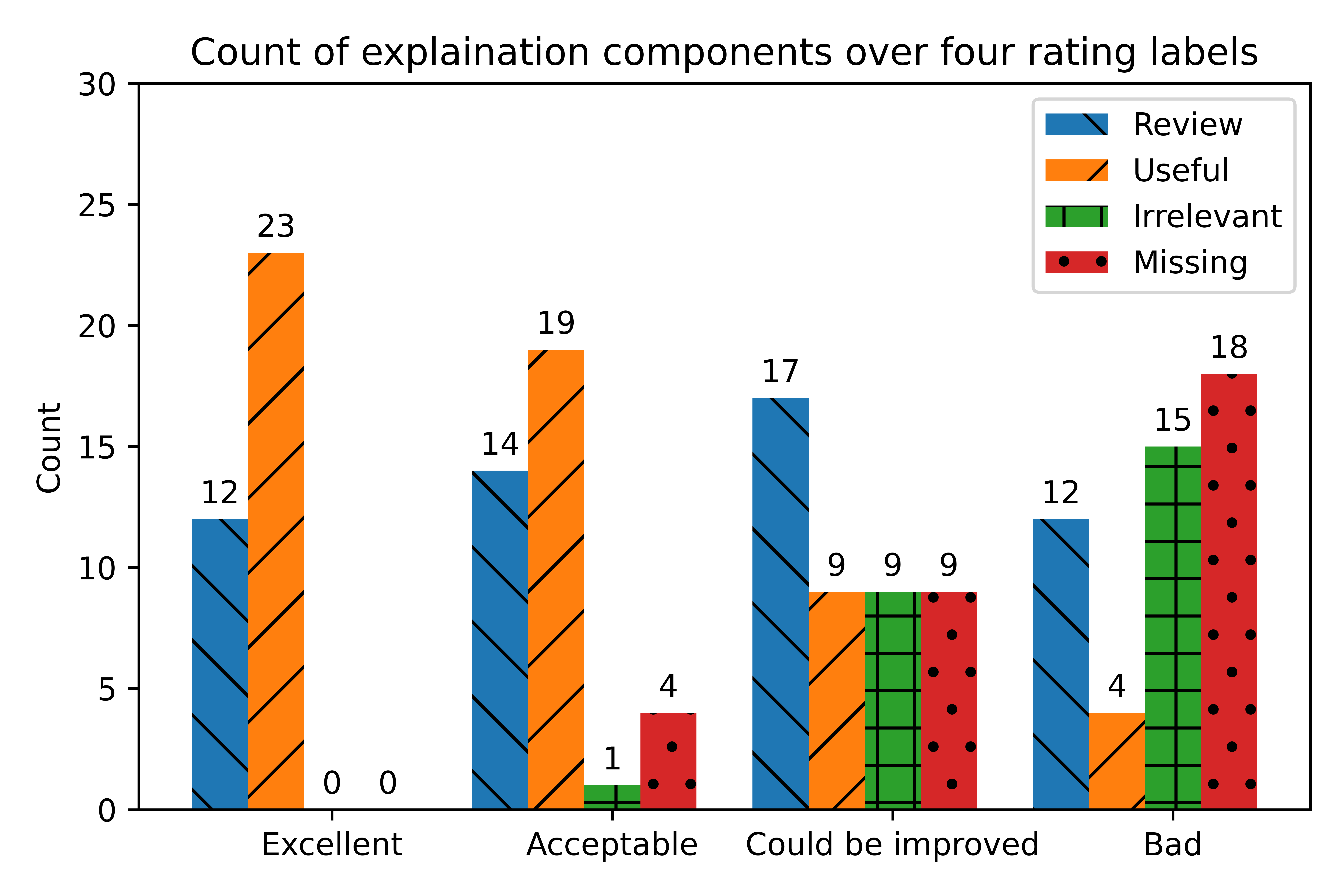}
    \caption{Distribution of component number in 100 natural language feedback of different rating labels.}
    \label{fig:my_label}
    \vspace{-\baselineskip}
\end{figure}
\vspace{-6pt}

\begin{table*}[ht!]
    \centering
    \resizebox{0.9\textwidth}{!}{
    \begin{tabular}{lp{15cm}}
        \toprule
        Rating label & Explanation \\
        \midrule
        Excellent & \hlc[lg]{This answers the question directly.} \hlc[lb]{This answer provides information and recommendation on how people and adolescent can protect themselves when going online during the Covid-19 pandemic.}  \\
        %Excellent  & \hlc[lo]{The answer addresses the question directly.} \hlc[lb]{It states that consumers should be able to get their money returned if a sports season is cancelled.} \\
        %Acceptable & \hlc[lb]{The answer is about testing for the two diseases,} \hlc[pink]{but does not discuss their interaction} \\
        Acceptable &  \hlc[lg]{This answer, while adequate, could give more information as}\hlc[ly]{this is a sparse answer for a bigger question of what one can do for elderly people during the pandemic.}\\
        Could be improved & \hlc[lg]{The answer relates and answers the question, but could improve grammatically and omit the "yes"}\\
        Could be improved & \hlc[lb]{The answer is about some of the online risks} \hlc[pink]{but not about how to protect against them.}\\
        %Bad & \hlc[pink]{This response doesn't answer the question of what symptoms are common to both viruses}. \hlc[ly]{The response describes a variety of things the flu and Covid-19 have in common: mortality rates, people who are at most risk for both, transmission rate/speed of transmission but doesn't mention anything about common symptoms.} \\
        Bad & \hlc[lg]{This does not answer the question.} \hlc[ly]{This information is about applying visa to work in critical sector.} \hlc[pink]{It does not provide any information on applying for Covid-19 pandemic visa event as asked in the question.}\\
        \bottomrule
    \end{tabular}
    }
    \caption{Examples of explanation and its associated rating label. Span color and their types of components: \hlc[lg]{generic and aspect review}; \hlc[lb]{summary of useful content}; \hlc[ly]{summary of irrelevant content}; \hlc[pink]{summary of missing content}}
    \label{tab:exp-examples}
\end{table*}

%\vspace{-1pt}
\section{Experimental Setup}

\feed contains two types of data.
One is pre-deployment data $\mathcal{D}_{\text{pre}}=(Q, A^+, \mathcal{A})$, where $Q$ is a question paired with its gold-standard answer passage $A^+$ from the domain corpus $\mathcal{A}$.
The other is post-deployment feedback data $\mathcal{D}_{\text{feed}}=(Q, A, Y, E)$, where $Q$ is a question paired with a candidate answer $A \in \mathcal{A}$ and corresponding feedback for the answer.
The feedback consists of a rating $Y$ and an explanation $E$.
We build two kinds of models on pre- and post-deployment data: 
RQA models on the pre-deployment data that can retrieve candidate answers for a given question, and feedback-enhanced RQA models on the post-deployment data that can rate an answer for a given question as well as generate an explanation for the answer.
We use this rating to rerank the answer candidates.
Therefore, in our setting, a feedback-enhanced RQA model is essentially a \textit{reranker}.
Keeping in mind the fact that real-world QA systems evolve quickly, we decouple the reranker model from the RQA model by using separate parameters for the reranker independent of the RQA model.
We train this reranker on the feedback data.
% training the reranker on the feedback data so that can rerank any given RQA predictions (and not just the predictions of the original deployed model).
This allows for the reranker to be reused across many RQA models. 
% In this work, we treat feedback-enhanced RQA models as a reranker.
% Further, we build \textit{feedback-enhanced} RQA models as follows: 1) we rerank the answer predictions using the ratings predicted by the feedback model; and 2) we generate explanations for each candidate answer. 
% The feedback models enhance RQA models in two ways: reranking the answer predictions using the ratings, and generating explanations for each answer.
% We call the resulting combination \textit{feedback-enhanced RQA} models.
% The feedback models are used as rerankers and for generating explanations to enhance an RQA model predictions: we call the resulting combination \textit{feedback-enhanced RQA} models.
We leave other ways to enhance RQA models with feedback data for future work.
Below, we describe the architectures for the RQA models and feedback-based rerankers.

\subsection{RQA Models (Pre-deployment)} 
\label{sec:baseline-models}
We use dense passage retrievers \cite{karpukhin2020dense} to build the RQA models, where the similarity between the question embedding and the passage embedding is used to rank candidates.
% In dense passage retrievers, question and passage embeddings are obtained using separate pretrained language models.
We use two variants of pre-trained models to obtain the embeddings: 1) BERT~\cite{devlin_bert_2019}, a pretrained Transformer encoder; and 2) BART~\cite{lewis_bart_2019}, a pretrained Transformer encoder-decoder.
For BERT, we use average pooling of token representations as the embedding, whereas for BART we use the decoder's final state.
While \citeauthor{karpukhin2020dense} use question-agnostic passage representations, we use a poly-encoder \cite{humeau_poly-encoders_2020} to build question-sensitive document representations.
In a poly-encoder, each passage is represented as multiple encodings, first independent of the question, but then a simple attention between the question and passage embeddings is used to compute question-sensitive passage representation, which is later used to compute the relevance of the passage for a given query.
\citeauthor{humeau_poly-encoders_2020} show that the poly-encoder architecture is superior to alternatives like the bi-encoder \cite{karpukhin2020dense} without much sacrifice in computational efficiency.\footnote{The performance results of poly-encoder and bi-encoder for our task are shown in \Cref{tab:valid-rslt}.}

Given pre-deployment training data $\mathcal{D}_{\text{pre}}=(Q, A^+, \mathcal{A})$, the RQA model parameterized by $\theta$ is trained to maximize the log-likelihood of the correct answer:
\begin{equation}\label{eqn:qa}
\begin{aligned}
    \mathcal{J}_{\theta} &= \log P_{\theta}(A^+|Q, \mathcal{A})\\
    P_{\theta}(A^i|Q,\mathcal{A}) &= \frac{\exp(S(Q, A^i))}{\sum_{A\in\mathcal{A}}\exp(S(Q, A))}
\end{aligned}
\end{equation}
Here $S(Q,A)$ denotes the dot product similarity between the question and passage embedding. %(we use dot product as the similarity function).\siva{What is the similarity function you used?} 
As it is inefficient to compute the denominator over all passages during training, we adopt an in-batch negative sampling technique~\citep{humeau_poly-encoders_2020}, merging all of the $A^+$ in the same minibatch into a set of candidates.

\subsection{Feedback-enhanced RQA models (Post-deployment)}
\label{sec:pipeline}

On the post-deployment data $\mathcal{D}_{\text{feed}}=(Q, A, Y, E)$, we train a reranker that assigns a rating to an answer and also generates an explanation.
We use BART parameterized by $\phi$ as the base of \exprate because it is ease to adapt it to both explanation generation and rating classification. 
The encoder of the BART model takes as input the concatenation $[Q; \text{SEP}; A]$, and the decoder generates an explanation $E$;
after that, an incremental fully-connected network predicts the rating $Y$ given the last hidden states of decoder. 
The rating is used to score QA pairs, whereas the generated explanation is passed to humans to make an informed decision of accepting the answer.
We also implement a variant where the model directly produces a rating without generating an explanation.
Since each candidate answer is annotated by different annotators, an answer could have multiple rating labels.
To account for this, we minimize the KL-divergence between the the target label distribution and the predicted distribution:
% We reshape the target distribution of ratings by simply normalizing its label occurrence and minimize the KL-divergence between the target distribution and the prediction distribution.
%
\begin{equation}
\begin{aligned}
    \mathcal{J}_{\phi'} = - D_{\text{KL}}(P(Y|Q, A)||P_{\phi}(Y|Q,A)),\\
    P(Y_i=y|Q_i,A_i) = \frac{C_{y, i}}
    {\sum_{y} C_{y, i}}
\end{aligned}
\end{equation}
where $C_{y, i}$ is the count of the rating label $y$ for the $i$-th feedback.

In order to enhance an RQA model with the reranker, we first select the top-$k$ candidates according to the RQA model (in practice we set $k=5$).
The reranker then takes as input the concatenation of the question and each candidate, then generates a rating for each answer. 
We simply sum up the scores from the RQA model and the reranker model. 
In practice, we found that using the reranker probability of \textit{excellent} worked better than normalizing the expectation of the rating score (from score 0 for label \textit{bad} to 3 for \textit{excellent}).
So, we score the candidate answers as follows:
\begin{equation}
    \begin{aligned}
    S(A|\mathcal{A}, Q) = &P_{\theta}(A=A^+|\mathcal{A}, Q)\\ &+ P_{\phi}(y=\textit{excellent}|A, Q)        
    \end{aligned}
\end{equation}

\section{Experiments and Results}

We organize the experiments based on the following research questions:
\vspace{-5pt}
\begin{itemize}[noitemsep,leftmargin=*]
\item RQ1: Does feedback data improve the base RQA model accuracy? 
\item RQ2: Does feedback data improve the accuracy of RQA models that are stronger than the base model?
\item RQ3: Do explanations aid humans in discerning between correct and incorrect answers?
\end{itemize}
\vspace{-6pt}
We answer these questions by comparing the RQA models with the feedback-enhanced RQA models.
The implementation and hyper-parameter details of each model are included in \Cref{sec:implementation-details}.

\begin{table*}[h]
    \centering
    \resizebox{0.8\textwidth}{!}{
    \begin{tabular}{l|ccccccc}
        \toprule
         Methods & Australia & US & Canada & UK & WHO & All & Beats\\
         \midrule
         %Pipeline models (Bi-encoder) &   &   &   &  &   & \\
         BERT RQA model \usym{2726} & 47.25	& 65.30	& 81.49 & 48.50 &	81.19 & 64.75 &  None \\
         %Pipeline models (Poly-encoder \& Reasoner \& Reweight) &   & 68.35  &  &  54.03 & 81.19 &  \\
         %Pipeline models (Poly-encoder \& Rater-KL) &   &  &  &  &  68 &  \\
         + \feedranker \usym{273B} & 55.13& 65.97 & 83.74 &	51.07 &	77.05	& 66.59 & \usym{2726} \usym{2746}\\ 
         + \vanillaranker \usym{2746} & 54.29 & 64.80 &	83.20 &			49.63 & 77.96 & 65.98 & \usym{2726}\\
          + \combinedranker \usym{2666} & \textbf{55.63} & \textbf{67.54} &	\textbf{84.99} & \textbf{53.21} & \textbf{78.51}	& \textbf{67.97} & \usym{2726} \usym{273B} \usym{2746}\\
         %MTL (Bi-encoder) &   &   \\ %\hline
         %\bartenc-MTL (Poly-encoder) & 51.39 &  \\
         %\bartenc-MTL (Poly-encoder \& Full) & 52.11 & 66.46  &  82.93 & 53.08 &   &   \\
         %MTL & \\
         %T5-MTL (Poly-encoder) & 37.17 & 48.42 & 70.92 & 36.02 & 69.31 &  \\
         \bottomrule
    \end{tabular}}
    \caption{Accuracy of the BERT RQA model, i,.e., the deployed model, and its enhanced variants on the test set. \feedranker is trained on the post-deployment feedback data, \vanillaranker is trained on the pre-deployment data and \combinedranker is trained on both. The column \textit{Beats} indicates that the model significantly outperforms ($p$-value $<0.05$) the competing methods. All of the results are averaged across 3 runs.}
    \label{tab:test-rslt-1}
    %\vspace*{-\baselineskip}
\end{table*}

\begin{table*}[h]
    \centering
    \resizebox{0.8\textwidth}{!}{
    \begin{tabular}{l|cccccccc}
        \toprule
         Methods & Australia & US & Canada & UK & WHO & All & Beats\\
         \midrule
         BART RQA model \usym{2648} & 52.88 & 68.47 & 82.49 & 51.29 & 81.97 & 67.42 &  None \\
         %Pipeline models (Poly-encoder \& Reasoner \& Reweight) &   & 68.35  &  &  54.03 & 81.19 &  \\
         %Pipeline models (Poly-encoder \& Rater-KL) &   &  &  &  &  68 &  \\
         + \feedranker \usym{2649} & 54.78 & 70.45 & 84.38 & 53.47 & 82.51 & 69.12 & \usym{2648} \usym{264A} \\ 
         + \textsc{VanillaReranker \usym{264A} } & 53.09 & 70.40 & 82.76 & 53.08 & 82.33 & 68.33 & \usym{2648} \\
         + \textsc{CombinedReranker \usym{1F340}} & \textbf{55.27} & \textbf{71.45} & \textbf{85.35} & \textbf{54.83} & \textbf{83.61} & \textbf{70.10} & \usym{2648} \usym{2649} \usym{264A}\\
         %MTL (Bi-encoder) &   &   \\ %\hline
         %\bartenc-MTL (Poly-encoder) & 51.39 &  \\
         %\bartenc-MTL (Poly-encoder \& Full) & 52.11 & 66.46  &  82.93 & 53.08 &   &   \\
         %MTL & \\
         %T5-MTL (Poly-encoder) & 37.17 & 48.42 & 70.92 & 36.02 & 69.31 &  \\
         \bottomrule
    \end{tabular}}
    \caption{Accuracy of the BART RQA model and its enhanced variants on the test set. Results are averaged across 3 runs.} %\textit{Beats} means that the model significantly outperforms ($p$-value $<0.05$) the competing methods.}
    \label{tab:test-rslt-2}
    \vspace*{-\baselineskip}
\end{table*}

\vspace{-2pt}
\subsection{RQ1: Does feedback data improve the base RQA model?}
\vspace{-2pt}
\paragraph{Model details.}
Our base model is a BERT RQA model which we deployed to collect feedback data to train the other models (\Cref{sec:baseline-models}).
%We train separate models for each domain.
%We deploy these models and collect feedback data.
%\Cref{tab:test-rslt-1} shows the results of the base QA model.

For the feedback-enhanced RQA model, we use the BART-based reranker described in \Cref{sec:pipeline}.
We train one single model for all domains.
We call this \feedranker. We compare two variants of \feedranker on validation set, one of which directly predicts the rating while the other first generates an explanation and then the rating. And we found the first one performs slightly better (Appendix~\Cref{tab:val-rslt-pipeline}).
We conjecture that learning an explanation-based rating model from the limited feedback data is a harder problem than directly learning a rating model.
Therefore, for this experiment, we only use the rating prediction model (but note that explanation-based rating model is already superior to the base RQA model).

To eliminate the confounding factor of having a larger number of model parameters introduced by the reranker, we train another reranker model on the pre-deployment data \vanillaranker and compare against the reranker trained on the feedback data.
To convert the pre-deployment data into the reranker's expected format, we consider a correct answer's rating label to be \textit{excellent}, and the randomly sampled answer candidates\footnote{We also tried using the top predictions from the base QA model, but found this approch leads to slightly worse performance than negative sampling.} to be \textit{bad}.
Note that this dataset is much larger than the feedback data.

Finally, we combine the training data of \feedranker and \vanillaranker and train the third reranker called \combinedranker.

To measure retrieval accuracy, we adopt Precision@1 (P@1) as our main metric.

\paragraph{Results.}
As shown in \Cref{tab:test-rslt-1}, the feedback-enhanced RQA model is significantly\footnote{We follow ~\citet{berg2012empirical} to conduct the statistical significant test} better than the base RQA model by 1.84 points. Although \vanillaranker improves upon the base model, it is weaker than \feedranker, and \combinedranker is a much stronger model than any of the models, indicating that learning signals presented in feedback data and the pre-deployment data are complementary to each other.
Moreover, we also see improved performance on the Canada domain, although feedback data was not collected for that domain.

%But is this improvement really due to the feedback data or is it because of the additional reranker model?
%To answer this question, we 

From these experiments, we conclude that feedback data can improve the accuracy of the base RQA model, not only for the domains for which feedback data is available but also for unseen domains (Canada).

\subsection{RQ2: Does feedback data improve the accuracy of RQA models that are stronger than the base model?}

If feedback data were only useful for the base RQA model, then its usefulness would be questionable, since the RQA development cycle is continuous and the base RQA model will eventually be replaced with a better model.
For example, we find that BART-based dense retriever is superior than the BERT RQA model:
\Cref{tab:valid-rslt} in Appendix E shows the results on validation set which indicate that BART RQA model overall performance is nearly 4 points better than the BERT RQA model.

To answer RQ2, we use the same \feedranker and \vanillaranker to rescore the BART RQA predictions, even though feedback data is not collected for this model.
We observe that the resulting model outperforms the BART RQA model in \Cref{tab:test-rslt-2}, indicating that the feedback data is still useful. Again, \feedranker is superior to \vanillaranker although the feedback data has fewer samples than the pre-deployment data, and the \combinedranker has the best performance.
%We retrain \vanillaranker on the pre-deployment data but with negative examples using the BART QA model, and the results have same pattern as for the base RQA model, i.e., \feedranker is superior to \vanillaranker although the feedback data has fewer samples than the pre-deployment data.We retrain the \combinedranker model and observe it improves the performance of \feedranker further.

These results suggest that the feedback data is useful not only for the base RQA model but also other stronger RQA models.

\subsection{RQ3: Do explanations aid humans in discerning between correct and incorrect answers?}
\label{ssec:human-eval}

We conduct a human evaluation to investigate whether explanations are useful from the perspective of users. 
Unfortunately, rigorous definitions and automatic metrics of explainability remain open research problems. 
In this work, we simulate a real-world scenario, where the user is presented an answer returned by the system as well as an explanation for the answer, and they are asked to determine whether the answer is acceptable or not.
\citet{goldberg2020explain} advocate utility metrics as proxies to measure the usefulness of explanations instead of directly evaluating an explanation since plausible explanations does not necessarily increase the utility of the resulting system.
Inspired by their findings, we measure if explanations can: 1) help users to make accurate decisions when judging an answer (with respect to a ground truth) and 2) improve the agreement among users in accepting/rejecting an answer candidate. 
The former measures the utility of an explanation and the latter measures if the explanations invoke the same behavioral pattern across different users irrespective of the utility of the explanation.
Note that agreement and utility are not tightly coupled. For example, agreement can be higher even if the utility of an explanation is lower when the explanation misleads end users to consistently select a wrong answer \cite{gonzalez-etal-2021-explanations,bansal2021does}.

\begin{table}[]
    %\centering
    \resizebox{0.9\linewidth}{!}{
    \begin{tabular}{l|cc}
        \toprule
        Explanation & Accuracy & Agreement\\ \midrule
         %BERT (Bi-encoder) & \\
         Blank &  69.17 & 0.31\\ \midrule
         Human-written & 88.33 & 0.80 \\
         BART feedback model & 81.67 & 0.71\\ \midrule
         BART summarization model & 74.17 & 0.30 \\
         \bottomrule
    \end{tabular}
    }
    \caption{Human evaluation results of the usefulness of explanations. Accuracy measures the utility of explanations in selecting the correct rating label for an answer, whereas agreement measures whether explanations invoke same behaviour pattern across users.}
    \label{tab:human_eval_rslt}
    \vspace{-1.5em}
\end{table}

We sample 60 feedback samples from the hidden split of the feedback data $\mathcal{D}_{\text{feed}}=(Q, A, Y, E)$ for evaluation purposes.\footnote{For simplicity, we merge the answer feedback labels \textit{good} and \textit{could be improved} into one label called \textit{partially correct}. The final set of answer labels we used are \textit{correct} (for original label \textit{excellent}), \textit{partially correct} and \textit{incorrect} (for original label \textit{bad}).}
% Since annotators deliberately thought and wrote an explanation about why an answer has certain rating label, the resulting data the resulting rating labels are ground-truth (The agreement between the ).
We evaluate four experimental setups on these samples which vary in the type of explanation shown to the end users: 1) no explanation; 2) human-written explanations; 3) explanations generated by the BART model trained on the feedback data (\Cref{sec:pipeline}); and 4) summary of the answer candidate generated by a strong fine-tuned BART-based summarization model.\footnote{\url{https://huggingface.co/facebook/bart-large-xsum}} 
The last setting is inspired from the observation in \Cref{sec:data-analysis} that a large portion of explanations contain summary of questions/answers. 
We investigate if conventional summary of an answer is as useful as an explanation.  %which is to distinguish the effect of explanation and other types of side information.
For each of these setups, two crowdworkers assign a rating label to each answer candidate indicating the quality of the answer.
Each setup has its own set of workers in order to avoid information-leakage across setups (this simulates A/B testing often used by production systems).

% We separate totally 240 QA pairs and assign them to four groups of human workers. 
% Given a QA pair, workers judge whether the answer candidate has successfully answered the question. 
% We again apply quality control methods and detail them in \Cref{sec:appendix-feedback}.

%\vspace*{-2mm}
\begin{table}[htp]
    \centering
    \resizebox{\linewidth}{!}{
    \small
    \begin{tabular}{p{\linewidth}}%p{6cm}}
        \toprule
        \midrule
        \textbf{Question: }  \textit{What are the restrictions to travelers landing in the US?}\\
        \midrule 
        \textbf{Answer: }  \textit{... Under current federal regulations, pilots must report all illnesses and deaths to CDC before arriving to a US destination. According to CDC disease protocols, if a sick traveler is considered to be a public health risk, CDC works with local and state health departments and international public health agencies to contact passengers and crew exposed to that sick traveler. Be sure to give the airline your current contact information when booking your ticket so you can be notified if you are exposed to a sick traveler on a flight. ...}\\
        \midrule
        \textbf{Generated explanation: }  \textit{This answer is irrelevant to the question. The question is asking about restrictions on travelers landing in the US and this answer provides information on what to do if there is a sick passenger on an international or domestic flight.}\\
        \midrule
        \textbf{Human explanation: }   \textit{This information does not answer the question. The question is asking for information on the restrictions to travelers landing in the US and this answer contains information on pilots reports on illnesses and deaths before arriving to a US destinations and airline notification when passengers are exposed to sick travelers on a flight.}\\
        \midrule
        \textbf{Generated Summary: } \textit{The Centers for Disease Control and Prevention (CDC) is working with airlines to help prevent illness and death on flights.}\\
        \midrule
        \midrule
        \textbf{Question: }\textit{Should I wear gloves when I go out in public to prevent Covid-19?}  \\
        \midrule
        \textbf{Answer: }  \textit{No, WHO does not recommend the regular use of gloves by people in the community. The wearing of gloves may increase risks, since it can lead to self-infection or transmission to others through touching contaminated surfaces which is not followed by removal of gloves and hand hygiene. Therefore, in public places such as supermarkets, in addition to physical distancing...}\\
        \midrule
        \textbf{Generated Explanation: }\textit{This answer is excellent because it unequivocally states that gloves are not recommended. Therefore, it answers the question directly. It goes on to explain why gloves aren’t recommended. This is a great answer.}\\
        \midrule
        %\textbf{Human explanation: } \textit{The question is answered directly.  The answer states not to wear gloves in public.}\\
        \textbf{Human explanation: } \textit{The question is answered directly.  The answer states not to wear gloves in public.}\\
        \midrule
        \textbf{Generated Summary: } \textit{The World Health Organization (WHO) has issued a guidance on the use of gloves in public places.}\\
        \midrule
        \midrule
        \textbf{Question: }\textit{Can students work in aged care?}\\
        \midrule
        \textbf{Answer:}\textit{...The Australian Government has announced that residential aged care facilities should take extra precautions to keep residents safe from COVID-19. Both individuals and management need to take responsibility for the health of visitors and staff at facilities to protect our most vulnerable community members. Aged care facilities will: closely monitor the health of staff...}\\
        \midrule
        \textbf{Generated Explanation: }\textit{This answer does not address whether or not students are allowed to work in aged care. This answer only provides information on precautions to be taken by aged care facilities to keep residents safe.}\\
        \midrule
        \textbf{Human Explanation: }\textit{The information here give explanation on guideline that aged care facility staffs should put in place and did not say anything about student working in aged care facility.}\\
        \midrule
        \textbf{Generated Summary: }\textit{Residents in aged care facilities across Australia are being urged to take extra precautions to prevent the spread of a deadly virus.}\\
        \midrule
        \bottomrule
    \end{tabular}}
    \caption{Examples of different explanation types: model-generated and human-written explanation and model-generated summary.}
    \label{tab:human-eval-expl}
\end{table}

We measure the workers' accuracy (average of the two workers) in determining the correctness of an answer with respect to the original annotation in \feed, as well as compute the agreement of workers with each other using Spearman correlation.
\Cref{tab:human_eval_rslt} presents the results.
All explanation types improve accuracy compared to the the model with no explanations.
This could be because any explanation forces the worker to think more about an answer.
The human-written explanations has the highest utility and also leads to the biggest agreement. Both the human-written explanations and the explanations generated by the BART feedback model have more utility and higher agreement than the BART summarization model.
In fact, the summarization model leads to lower agreement. %than showing no explanation.

These results indicate that explanations based on feedback data are useful for end users in discerning correct and incorrect answers, and they also improve the agreement across users.

% While the generated summary helps increase the accuracy, it fails to make a difference in terms of users' agreement. We also notice that there is a large margin between the human-written feedback and model-generated feedback indicating ample room for explanation generation. 
Table~\ref{tab:human-eval-expl} shows some examples of explanation that helps the users make more informed and accurate decision. 
In the first example, the model-generated explanation points out the gap between the question and the answer candidate, though there are a large number of overlapping keywords.
Meanwhile, human explanations are generally more abstractive and shorter in nature (e.g., see the second example).

\vspace{-3pt}
\section{Related work}
\vspace{-3pt}
\label{sec:related_works}
\paragraph{Retrieval-based question answering} has been widely studied, from early work on rule-based systems~\cite{kwok2001scaling}, to recently proposed neural-based models~\cite{yang2019BERTserini, karpukhin2020dense}. 
Most existing work focuses on improving the accuracy and efficacy by modification of a neural architecture~\cite{karpukhin2020dense, humeau_poly-encoders_2020}, incorporation of external knowledge~\cite{ferrucci2010watson}, and retrieval strategy~\cite{kratzwald2018adaptive}. 
These methods focus on the pre-deployment stage of RQA models.

By contrast, we investigate methods to improve a RQA model post-deployment with interactive feedback. The proposed methods are agnostic to the architecture design and training methods of the base RQA model.
\vspace{-3pt}
\paragraph{Learning from user feedback} has been a long standing problem in natural language processing. 
Whilst earlier work proposes methods for using implicit feedback---for instance, using click-through data for document ranking \citep{Joachims2002opt}---recent work has explored explicit feedback such as explanations of incorrect responses by chatbots~\cite{li2016dialogue, weston2016dialogbased} and correctness labels in conversational question answering and text classification~\cite{campos2020improving}. 
However, the feedback in these studies is automatically generated using heuristics, whereas our feedback data is collected from human users. 
\citet{hancock2019learning} collect suggested responses from users to improve a chatbot, while we investigate the effect of natural feedback for RQA models. %suggested response for chatbots~\citep{, hancock2019learning, sreedhar-etal-2020-learning}.

\vspace{-3pt}
\paragraph{Explainability and Interpretability} has received increasing attention in the NLP community recently. 
This paper can be aligned to recent efforts in collecting and harnessing explanation data for language understanding and reasoning tasks, such as natural language inference \cite{camburu_e-snli_2018, kumar_nile_2020}, commonsense question answering~\cite{rajani2019explain}, document classification~\cite{srivastava2017joint}, relation classification~\cite{murty2020expbert}, reading comprehension~\cite{lamm2020qed}, and fact checking~\cite{alhindi2018liarplus}. 
The type of feedback in \feed differs from the existing work in several aspects: 1) \feed has feedback data for both positive and negative examples, while most of other datasets only contains explanations of positive ones; 2) \feed has both structured and unstructured feedback, while previous work mainly focuses on one of them; 3) The feedback in \feed is collected post-deployment; 4) While previous work aims to help users interpret model decisions, we investigate whether feedback-based explanations increase the utility of the deployed system.

% since it is most understandable from the perspective of users. 

%Open-domain question answering is a classic "machine-learning at scale" problem where the model is tasked with understanding questions posed by the user, retrieving and ranking relevant documents from a large corpus and utilizing the retrieved subset of textual data to formulate appropriate answilers to the query. Recent developments in deep learning has resulted in neural models for information retrieval showing tremendous success. 
\vspace{-3pt}
\section{Conclusion}
\vspace{-3pt}
In this work, we investigate the usefulness of feedback data in retrieval-based question answering. We collect a new dataset \feed{}, which contains interactive feedback in the form of ratings and natural language explanations.
We propose a method to improve the RQA model with the feedback data, training a reranker to select an answer candidate as well as generate the explanation. 
We find that this approach not only increases the accuracy of the deployed model but also other stronger models for which feedback data is not collected. 
Moreover, our human evaluation results show that both human-written and model-generated explanations help users to make informed and accurate decisions about whether to accept an answer.

\section{Limitations and Ethical consideration}
The training and inference of a reranker with feedback data increases the usage of computational resources.
We note that our feedback collection setup is a simulation of a deployed model. 
The feedback in real-world systems may contain sensitive information that should be handled with care.
Moreover, real-world feedback could be noisy and is prone to adversarial attacks.
% We also notice that there is a potential privacy issue when collecting feedback data from users without their approval.
% In this work, we eliminate this issue by recruiting crowd-workers to provide feedback. 

\section{Acknowledgements} 
We would like to thank Andreas Madsen, Nathan Schucher, Nick Meade and Makesh Narsimhan for their discussion and feedback on our manuscript.
We would also like to thank the Mila Applied Research team, especially Joumana Ghosn, Mirko Bronzi, Jeremy Pinto, and Cem Subakan whose initial work on the Covid-19 chatbot inspired this work.
%The \feed dataset is partly funded through the Scale AI program for tackling the Covid-19 pandemic through collaboration between Mila and Dialogue.
This work is funded by Samsung Electronics. JC and SR acknowledge the support of the NSERC Discovery Grant program and the Canada CIFAR AI Chair program.
The computational resource for this project is partly supported by Compute Canada.
\bibliography{ref}
\bibliographystyle{acl_natbib}

\newpage
%~\newpage
%\hfill
\appendix
\section{Details of Data Collection}
\label{sec:appendix-data-collect}
\paragraph{Passage curating}
After we scraped the websites, we collect the questions and answers in the Frequently-Asked-Questions pages directly. For those pages without explicit questions and answers, we extract the text content as passages and proceed to question collection.

\paragraph{Question collection}
We hire crowd-source workers from English-speaking countries at the Amazon MTurk platform to write questions conditioned on the extracted passages. The workers are instructed not to ask too generic questions or copy and paste directly from the passages.

A qualification test with two sections is done to pick up the best performing workers. In the first section, the workers are asked to distinguish the good question from the bad ones for given passages. The correct and incorrect questions were carefully designed to test various aspects of low-quality submissions we had received in the demo run.
The second section is that writing a question given a passage. We manually review and score the questions. 
We paid 0.2\$ to workers for each question.

\section{Details of Feedback Collection}
\label{sec:appendix-feedback}
We asked the workers to provide rating and natural language feedback for question-answer pairs. For qualification test, we labeled the rating for multiple pairs of questions and answers. The workers are selected based on their accuracy of rating labeling. 
We paid 0.4\$ to workers for each feedback.

\section{Details of Human Evaluation}
\label{sec:appendix-human-anno}

The worker assignment is done to make sure a worker rates the same question-answer pair only once. Otherwise there is risk that the workers just blindly give the same judgement for a certain QA pair.

We adopt the qualification test similar to the one for feedback collection. We also include some dummy QA pairs, whose answer candidate were randomly sampled from the corpora, and we filter out the workers who fail to recognize them.
We paid 0.3\$ to workers for each QA pair.

\section{Implementation Details}
\label{sec:implementation-details}
Throughout the experiments, we have used 4 32-GB Nvidia Tesla V100. The hyperparameter (learning rate, dropout rate) optimisation is performed for the RQA models only and standard fine-tuning hyperparameters of BART are used for building the \feedranker model. We set batch size as $16$. We truncate the questions and passages to 50 and 512 tokens, respectively. The models are trained with 40 epochs. 
For our hyperparameter search, we have used 5 trials and while reporting the final results the best hyperparameter variant's performance was averaged across 3 different runs. All experiment runs were finished within 20 hours.
\begin{table}[t!]
    \centering
    \resizebox{\linewidth}{!}{
    \begin{tabular}{lrrrrr}
        \toprule
         & $lr$
         & Dropout
         % & Pooling operation
          \\ \midrule
         BERT (Bi-encoder) & 5.0e-05 & 0.1 \\ % & average \\ 
         BERT (Poly-encoder) & 5.0e-05 & 0.1 \\ % & average \\
         BART (Bi-encoder) & 9.53e-05 & 0.01026 \\ % & EOS \\ %\hline
         BART (Poly-encoder) & 4.34e-05 & 0.1859 \\ % & EOS\\
         \feedranker & 5.0e-05 & 0.1 \\ % & / \\
         \bottomrule
    \end{tabular}
    }
    \caption{Hyper-parameter setting of different variants of QA models as well as \exprate and \rate. There is no pooling operation in the latter two models.}
    \label{tab:hypers}
\end{table}

\section{Validation performance}
%\FloatBarrier
\begin{table*}[t]
    %\vspace*{0pt}
    \centering
    % \resizebox{\linewidth}{!}{
    %\vspace*{-20pt}
    \begin{tabular}{l|cccccc}
        \toprule
         Methods & Australia & US & Canada & UK & WHO & All\\ \midrule
         BERT (Bi-encoder) & 44.57 & 64.24 & 81.12 & 50.55 & 81.85	& 64.47 \\
         BERT (Poly-encoder) & 47.25 & 65.30 & 81.49 & 48.50 & 81.19 & 	64.75 \\
         %BERT (Vanilla Poly-encoder) & 46.87 & 63.61 & 81.40 & 51.11 & 80.53 & 64.70 \\
         %BART-BASE-encoder+decoder (Bi-encoder) & 53.11  & 68.10  & 82.23  & 58.25  & 82.85  & 68.91  \\
         %BART-BASE-encoder+decoder (Poly-encoder) & 54.31  & 68.29  & 82.42  & 59.09  & 81.92  & 69.20  \\
         %\midrule
         %\multicolumn{6}{c}{BART-QA}\\ \midrule
         BART (Bi-encoder) & 47.13 & 67.62 & 86.01 & 55.06 & 85.48 & 68.26 \\
         BART (Poly-encoder) & 49.17 & 66.98 & 85.75 &	54.27 & 87.46 & 68.73 \\
         %\bartenc (Bi-encoder) &  \\
         %\midrule
         %\multicolumn{6}{c}{Post-deployment}\\ \midrule
         %\textsc{Pipeline-Rate} & 51.09  & 68.57  & \textbf{86.84}  & \textbf{58.21}  & \textbf{88.78} & \textbf{70.70} \\
         \bottomrule
    \end{tabular}
    %}
    \caption{The accuracy of different RQA models on the validation set. All of the results are averaged across 3 runs.}
    \label{tab:valid-rslt}
\end{table*}

\begin{table*}[t]
    \centering
    %\vspace{-10pt}
    % \resizebox{\linewidth}{!}{
    %\vspace*{-10pt}
    \begin{tabular}{p{12em}|ccccccc}
        \toprule
         Methods & Australia & US & Canada & UK & WHO & All\\
         \midrule
         %Pipeline models (Bi-encoder) &   &   &   &  &   & \\
         \multicolumn{7}{c}{BART RQA model}\\ \midrule
         %Pipeline models (Poly-encoder \& Reasoner \& Reweight) &   & 68.35  &  &  54.03 & 81.19 &  \\
         %Pipeline models (Poly-encoder \& Rater-KL) &   &  &  &  &  68 &  \\
         BART RQA model & 49.17 & 66.98 & 85.75 &	54.27 & 87.46 & 68.73 \\
         + \feedranker with explanation-based rating & 51.34 & 69.09 & 84.20 & 56.87 & 87.79 & 69.86\\
         + \feedranker with rating only & 51.09	& 68.57 & 86.84	& 58.21 & 88.78 & 70.70\\
         \midrule
         \multicolumn{7}{c}{BERT RQA model}\\ \midrule
         % BERT RQA model & 46.87 & 63.61 & 81.40 & 51.11 & 80.53 & 64.70\\
         BERT RQA model & 47.25 & 65.30 & 81.49 &	48.50 & 81.19 &	64.75 \\
         + \feedranker with explanation-based rating & 51.34 &	70.15 & 83.72 & 53.71 &	84.49 &	68.68 \\
         + \feedranker with rating only & 51.09 & 68.46 &	84.18 & 55.69 &	85.15 & 68.91 \\
         %Pipeline models (Poly-encoder \& Reasoner \& Reweight) &   & 68.35  &  &  54.03 & 81.19 &  \\
         %Pipeline models (Poly-encoder \& Rater-KL) &   &  &  &  &  68 &  \\
         %MTL (Bi-encoder) &   &   \\ %\hline
         %\bartenc-MTL (Poly-encoder) & 51.39 &  \\
         %\bartenc-MTL (Poly-encoder \& Full) & 52.11 & 66.46  &  82.93 & 53.08 &   &   \\
         %MTL & \\
         %T5-MTL (Poly-encoder) & 37.17 & 48.42 & 70.92 & 36.02 & 69.31 &  \\
         \bottomrule
    \end{tabular}
    %}
    \caption{Accuracy of \textsc{Pipeline} models using different feedback data to train the re-ranker on the validation set. All of the results are averaged across 3 runs.} %\textit{Beats} means that the model significantly outperforms ($p$-value $<0.05$) the competing methods.}
    \label{tab:val-rslt-pipeline}
    %\vspace*{-\baselineskip}
\end{table*}
In addition to the Poly-encoders, we also explore Bi-encoder and we have found that its performance is consistently worse. \Cref{tab:valid-rslt} presents the performance of base QA models with different pre-trained Transformer models and encoding methods on the validation set.

\newpage
\vspace*{-2pt}

\vfill

\end{document}